\documentclass{article}

\usepackage[preprint]{neurips_2026}

\usepackage[utf8]{inputenc} 
\usepackage[T1]{fontenc}    
\usepackage{hyperref}       
\usepackage{url}            
\usepackage{booktabs}       
\usepackage{amsfonts}       
\usepackage{nicefrac}       
\usepackage{microtype}      
\usepackage{xcolor}         
\usepackage{amsmath}
\usepackage{graphicx}
\usepackage{multirow}
\usepackage{algorithm}
\usepackage{algpseudocode}

\title{STARRY: Spatial-Temporal Action-Centric World Modeling for Robotic Manipulation}

\author{%
  Yuxuan Tian\textsuperscript{1,2} \quad
  Yurun Jin\textsuperscript{3,4} \quad
  Bin Yu\textsuperscript{2,5} \quad
  Yukun Shi\textsuperscript{3} \quad
  Hao Wu\textsuperscript{2,6}\\
  \textbf{Chi Harold Liu\textsuperscript{1 \thanks{Corresponding authors}}} \quad
  \textbf{Kai Chen\textsuperscript{2,3,7 \footnotemark[1] \ \thanks{Project leader}}} \quad
  \textbf{Cong Huang\textsuperscript{2,3 \footnotemark[1]}} \quad
  \\[2ex]
  \textsuperscript{1}Beijing Institute of Technology \qquad
  \textsuperscript{2}Zhongguancun Academy\\
  \textsuperscript{3}Zhongguancun Institute of Artificial Intelligence\\
  \textsuperscript{4}University of Science and Technology of China \\
  \textsuperscript{5}Harbin Institute of Technology \quad
  \textsuperscript{6}East China Normal University \quad
  \textsuperscript{7}DeepCybo
}

\begin{document}

\maketitle

\begin{abstract}
  Robotic manipulation requires reasoning about future spatial-temporal interactions and geometric constraints, yet existing Vision-Language-Action (VLA) policies often leave predictive representation weakly coupled with action execution, causing failures in tasks requiring precise spatial-temporal coordination. We propose \textbf{STARRY}, a world-model-enhanced action-generation policy that aligns spatial-temporal prediction and action generation by jointly denoising future spatial-temporal latents and actions through a unified diffusion process. To bridge 2D visual tokens and 3D metric control, STARRY introduces Geometry-Aware Selective Attention Modulation (GASAM), which converts predicted depth and end-effector geometry into token-aligned weights for selective action-attention modulation. On RoboTwin 2.0, STARRY achieves \textbf{93.82\%} / \textbf{93.30\%} average success under Clean and Randomized settings across 50 bimanual tasks. Real-world experiments show that STARRY improves average success from \textbf{42.5\%} to \textbf{70.8\%} compared with $\pi_{0.5}$. These results demonstrate the effectiveness of action-centric spatial-temporal world modeling for spatially and temporally demanding robotic manipulation.
\end{abstract}

\section{Introduction}

Vision-Language-Action (VLA) models have emerged as a prominent paradigm for general-purpose embodied agents by unifying perception, language, and action~\cite{black2025pi0,black2025pi05,din2025vlareview,rt2}. Although large-scale vision-language pretraining provides strong semantic understanding and instruction following, many VLA policies remain \emph{reactive}, primarily relying on current or short-history observations without explicitly modeling future robot-object interaction states~\cite{lv2025f1}. This limits language-conditioned manipulation, where tasks such as hanging a mug, handover, or container placement require anticipating object geometry, contact regions, and end-effector trajectories. Errors in these local relations can cause unstable grasps, collisions, or failed placements.

Recent work incorporates world models into policy learning by predicting future observations or latent video states, leading to \emph{world-model-enhanced action-generation policies}~\cite{bi2025motus, internvla2026, zhu2025wmpo, cen2025worldvla, gigaworld2025}. Nevertheless, future prediction alone does not necessarily yield better control: visually plausible futures may still fail to expose action-relevant spatial constraints such as object handles, contact surfaces, openings, obstacles, and the neighborhood around the end effector.

We identify two common limitations in many existing predictive manipulation policies. First, their future representations are often optimized for perceptual or temporal consistency rather than action relevance, creating a mismatch between prediction and control~\cite{li2025worldmodelsurvey, hy2026embodied}. Second, spatial information is usually fused through globally shared representations, making it difficult to distinguish decision-critical regions from background context~\cite{cen2025worldvla, zhen2025learning4d, upvla}. As a result, the predictive capability of world models may not be fully converted into effective action generation, especially in contact-rich or spatially constrained tasks where local spatial-temporal relations matter.

Our key insight is that world models for manipulation should be both \emph{action-centric} and \emph{geometry-grounded}: they should predict not only how the scene evolves, but also where future interactions are critical for action generation. Based on this insight, we propose \textbf{STARRY} (\textbf{S}patial-\textbf{T}emporal \textbf{A}ction-centric \textbf{R}epresentation and \textbf{R}easoning Polic\textbf{Y}), a \textbf{world-model-enhanced action-generation policy} for robotic manipulation. STARRY jointly denoises future spatial-temporal latent variables and action sequences within a unified generative policy, aligning future prediction with action generation. To further incorporate geometric guidance, we introduce \textbf{Geometry-Aware Selective Attention Modulation} (GASAM), which uses predicted geometry to emphasize action-relevant visual tokens in the action attention branch while preserving video modeling and semantic understanding.

\textbf{Our contributions are summarized as follows:}
\begin{itemize}
\item We propose \textbf{STARRY}, a world-model-enhanced action-generation policy that jointly models future spatial-temporal dynamics and action sequences, enabling action-centric spatial-temporal modeling for language-conditioned robotic manipulation.
\item We introduce \textbf{Geometry-Aware Selective Attention Modulation} (GASAM), which converts predicted depth and end-effector geometry into token-aligned weights and selectively injects them into the action attention branch to emphasize decision-critical regions.
\item We validate STARRY through simulation and real-world experiments. On RoboTwin 2.0, it achieves \textbf{93.82\% / 93.30\%} average success under Clean and Randomized settings, while real-world experiments and ablations confirm the benefits of spatial-temporal modeling and GASAM.
\end{itemize}

\section{Related work}

\subsection{Vision-Language-Action policies}

Vision-Language-Action (VLA) models unify perception, language, and action, forming a dominant paradigm for embodied manipulation~\cite{black2025pi05, din2025vlareview}. Representative works such as RT-2 and $\pi_{0.5}$ leverage large-scale vision-language pretraining to learn rich semantic representations, enabling direct mapping from multi-modal inputs to action outputs~\cite{rt1,rt2,saycan,vima,black2025pi0,black2025pi05}. However, most VLA policies remain \emph{reactive}, mapping observations directly to actions without explicitly modeling future states, limiting their ability to leverage temporal foresight for action generation~\cite{lv2025f1}. In addition, existing methods are largely 2D-centric and lack explicit geometric reasoning, relying primarily on appearance-based features, which restricts performance in spatially constrained tasks~\cite{geopredict}.

\subsection{World models and predictive learning}

World models learn environment dynamics to support prediction and decision-making~\cite{li2025worldmodelsurvey}. Recent advances span latent predictive learning (e.g., JEPA), which focuses on learning temporally consistent representations, and large-scale generative models such as DreamDojo and WoW that model future observations from video or interaction data~\cite{vjepa, dreamdojo, wow, gigaworld2025}. Unified architectures further couple video and action modeling for scalable policy learning. Despite strong predictive capabilities, these models often exhibit a gap between prediction quality and control utility~\cite{worldarena}, as they primarily optimize perceptual fidelity rather than action relevance, and emphasize temporal dynamics with limited modeling of spatial geometry.

\subsection{World-model-enhanced action generation}

Recent work integrates world models into action generation by leveraging predicted futures for policy learning. Representative approaches include foresight-based action prediction (e.g., F1), which formulates action generation as a prediction-guided inverse dynamics problem, and unified video-action modeling frameworks such as Motus that jointly learn latent representations for perception and control~\cite{lv2025f1, bi2025motus, uwm}. While these methods improve temporal consistency and multi-modal integration, they provide limited modeling of spatial geometry and lack mechanisms for selectively emphasizing action-relevant regions. 

Our work addresses the above gap by learning action-centric spatial-temporal representations and introducing geometry-aware selective modulation for more effective action generation~\cite{act,diffusionpolicy}.

\begin{figure*}[t]
\centering
\includegraphics[width=\textwidth]{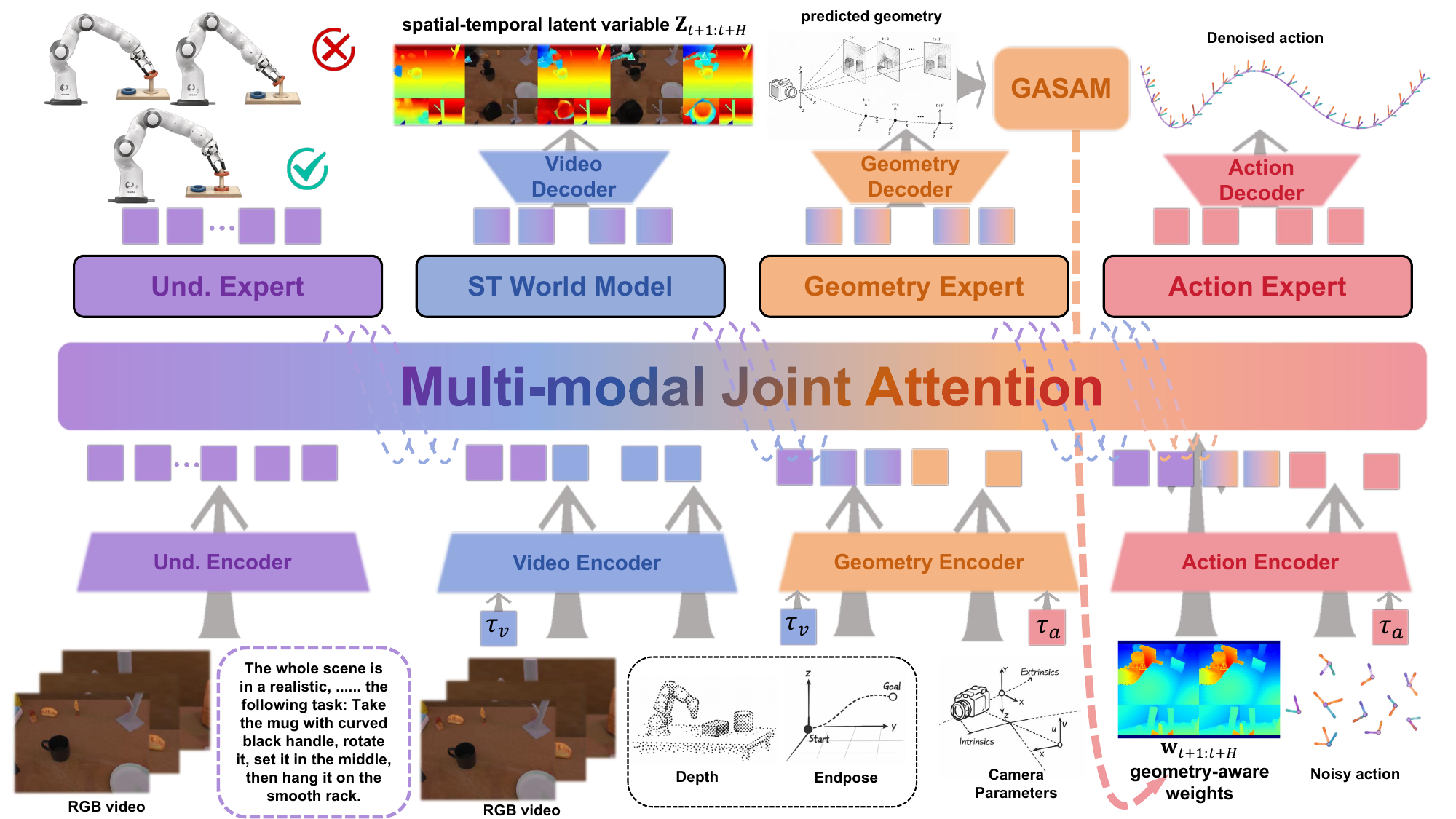}
\caption{
Overview of \textbf{STARRY}. The policy contains four modules: \emph{Understanding Expert}, \emph{Spatial-Temporal (ST) World Model}, \emph{Geometry Expert}, and \emph{Action Expert}. The \emph{ST World Model} predicts future spatial-temporal latent variables, while the \emph{Geometry Expert} and GASAM generate geometry-aware weights to selectively modulate the action branch.
}
\label{fig:framework}
\end{figure*}

\section{STARRY: Spatial-Temporal Action-Centric Representation and Reasoning}

We propose \textbf{STARRY}, a world-model-enhanced action-generation policy for embodied manipulation that jointly models future temporal dynamics and spatial geometry for action generation. We first present the overall architecture (Sec.~\ref{section3.1}), followed by the \emph{ST World Model} (Sec.~\ref{section3.2}), the Geometry-Aware Selective Attention Modulation mechanism (Sec.~\ref{section3.3}), and the model training and data construction (Sec.~\ref{section3.4}).

\subsection{Architecture overview}
\label{section3.1}

\paragraph{Problem formulation.}
We consider language-conditioned embodied manipulation, where the agent observes
$\mathbf{o}_t = \{\mathbf{I}_t, \mathbf{D}_t, \mathbf{c}_t, \mathbf{p}_t, \mathbf{l}\}$
at time step $t$, including multi-view RGB-D observations $\{\mathbf{I}_t, \mathbf{D}_t\}$, camera parameters $\mathbf{c}_t$, current pose $\mathbf{p}_t$, and the language instruction $\mathbf{l}$. The goal is to generate a future action sequence $\mathbf{a}_{t+1:t+H}$. 
Instead of directly mapping $\mathbf{o}_t$ to actions, STARRY introduces two internal structures: a future spatial-temporal latent variable $\mathbf{z}_{t+1:t+H}$ and geometry-aware modulation weights $\mathbf{w}_{t+1:t+H}$. The future latents and actions are jointly modeled as
\begin{equation}
\pi_\theta\!\left(
\mathbf{a}_{t+1:t+H}, \mathbf{z}_{t+1:t+H}
\mid \mathbf{o}_t
\right).
\end{equation}
Here, $\mathbf{z}_{t+1:t+H}$ captures predictive spatial-temporal structure aligned with action generation, while $\mathbf{w}_{t+1:t+H}$ is generated from predicted future geometry and used to inject action-relevant spatial constraints into the action branch.

\paragraph{Architecture.}

As shown in Fig.~\ref{fig:framework}, STARRY consists of four modules: the \emph{Understanding Expert}, \emph{Spatial-Temporal (ST) World Model}, \emph{Geometry Expert}, and \emph{Action Expert}. The core design is that future spatial-temporal latents and actions are jointly denoised over the same horizon, while geometry-aware weights are selectively applied only to the action attention branch.

Concretely, the \emph{Understanding Expert} provides semantic grounding from visual-language inputs, while the \emph{ST World Model} and \emph{Action Expert} jointly denoise future spatial-temporal latents $\mathbf{z}_{t+1:t+H}$ and actions $\mathbf{a}_{t+1:t+H}$ through Multi-modal Joint Attention. In parallel, the \emph{Geometry Expert} predicts future geometric states, which GASAM converts into token-aligned weights $\mathbf{w}_{t+1:t+H}$ to selectively modulate the action attention branch. This enables action-relevant spatial reasoning while preserving video modeling and visual-language understanding.

\begin{figure}[t]
\centering
\includegraphics[width=0.75\linewidth]{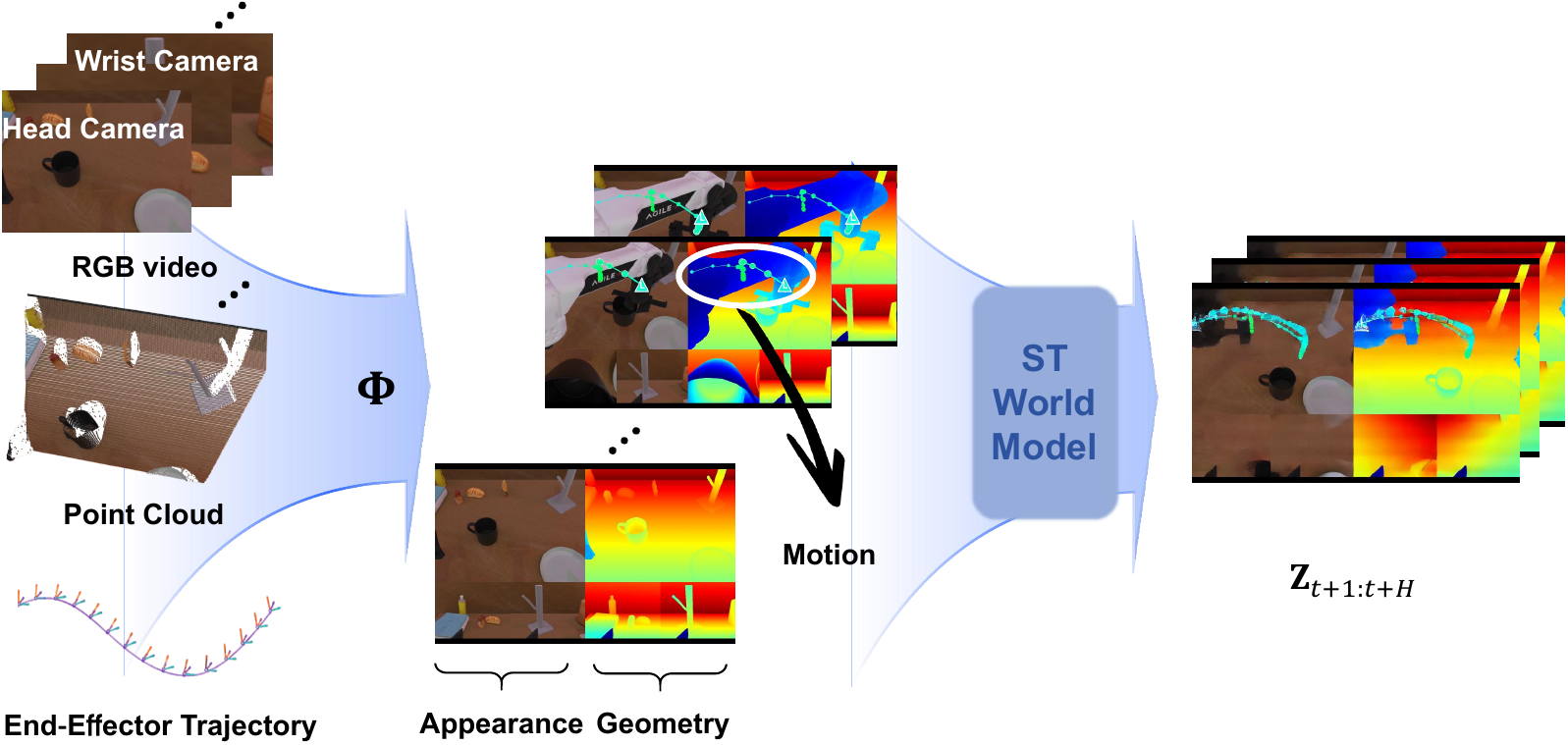}
\caption{
Construction of spatial-temporal inputs for the \emph{ST World Model}. Multi-modal signals, including RGB observations, depth observations, and end-effector trajectories, are unified into a shared representation capturing appearance, geometry, and motion, which is then used for future spatial-temporal prediction.
}
\label{fig:st_world_model}
\end{figure}

\subsection{ST World Model}
\label{section3.2}

Existing world-model-enhanced policies typically use future observation or latent video prediction to improve temporal coherence in action generation. However, appearance-level or temporally consistent futures do not necessarily encode the spatial relations that determine manipulation success, such as end-effector motion, object geometry, and contact-relevant regions. The \emph{Spatial-Temporal (ST) World Model} addresses this by learning a unified future spatial-temporal latent variable $\mathbf{z}_{t+1:t+H}$ that integrates appearance, motion, and geometry, as shown in Fig.~\ref{fig:st_world_model}. This representation provides structured future information that is directly shared with action generation.



Given observations within a temporal window $[t_0,t]$, we construct a unified spatial-temporal representation from multi-view RGB images $\{\mathbf{I}_t^{c}\}_{c}$, depth observations $\{\mathbf{D}_t^{c}\}_{c}$, and 3D end-effector trajectories $\{\mathbf{e}_{\tau}^{m}\}_{\tau \le t,m}$. We first project trajectories into each camera view using camera intrinsics $K$ and extrinsics $T$: $\mathcal{C}_t^{c}=\{K_t^{c},T_t^{c}\}$, and then compose appearance, geometry, and motion into a unified representation:
\begin{equation}
\mathbf{u}_{\tau}^{c,m}
=
\Pi\!\left(K_t^{c},T_t^{c},\mathbf{e}_{\tau}^{m}\right),
\qquad
\mathbf{x}_t
=
\Phi\!\left(
\{\mathbf{I}_t^{c}\}_{c},
\{\mathbf{D}_t^{c}\}_{c},
\{\mathbf{u}_{\tau}^{c,m}\}_{\tau \le t,c,m}
\right).
\end{equation}
Here, $\Pi(\cdot)$ denotes camera projection, and $\Phi(\cdot)$ denotes the spatial-temporal composition function, which arranges the multi-view RGB frames, depth observations, and projected trajectories into a fixed RGB-D layout. The resulting sequence $\mathbf{X}_{t_0:t}=\{\mathbf{x}_{t_0},\dots,\mathbf{x}_t\}$ provides a structured description of past spatial-temporal states.

We encode $\mathbf{X}_{t_0:t}$ into video tokens $\mathbf{v}_{t_0:t}$ and combine them with historical actions $\mathbf{a}_{\le t}$ to predict future latent variables:
\begin{equation}
\mathbf{z}_{t+1:t+H} = f_{\theta}^{\text{ST}}\!\left(\mathbf{v}_{t_0:t}, \mathbf{a}_{\le t}\right).
\end{equation}
Here, $f_{\theta}^{\text{ST}}$ is implemented as a diffusion-based model. By explicitly integrating appearance, trajectory, and geometry in the input representation, $\mathbf{z}_{t+1:t+H}$ captures structured spatial-temporal information, including scene evolution, end-effector motion, and geometric constraints, providing effective conditioning for action generation.

\subsection{Geometry-Aware Selective Attention Modulation}
\label{section3.3}

Although the \emph{ST World Model} provides future spatial-temporal latents, action generation still requires explicit alignment between 2D visual tokens and 3D physical control. Standard joint attention matches action tokens with 2D visual tokens through query-key dot products in token space, forcing the policy to infer perspective and depth relations implicitly. This may cause large 3D control deviations when visual similarity does not reflect metric distance. To bridge this gap, we propose \emph{Geometry-Aware Selective Attention Modulation} (GASAM), which uses predicted depth and end-effector geometry to lift 2D visual tokens into metric 3D space, thereby explicitly aligning visual observations with physical control.


\paragraph{Geometry prediction.}


While future latent variables $\mathbf{z}_{t+1:t+H}$ capture future spatial-temporal structure, they remain latent and noise-dependent during denoising, making them unsuitable for direct geometric operations such as depth unprojection or end-effector--scene distance computation. To obtain stable and interpretable spatial guidance, we introduce a \emph{Geometry Expert}. Conditioned on $\mathbf{o}_t$, video tokens $\mathbf{v}_{t_0:t}$, historical actions $\mathbf{a}_{\le t}$, and diffusion timesteps $(\tau_v,\tau_a)$ that indicate the noise levels of the video and action branches, it predicts future depth sequences $\hat{\mathbf{D}}_{t+1:t+H}$ and end-effector positions $\hat{\mathbf{p}}_{t+1:t+H}$:
\begin{equation}
(\hat{\mathbf{D}}_{t+1:t+H}, \hat{\mathbf{p}}_{t+1:t+H})
=
g_{\phi}\!\left(\mathbf{v}_{t_0:t}, \mathbf{a}_{\le t}, \mathbf{o}_t, \tau_v, \tau_a\right),
\end{equation}
where $g_{\phi}$ denotes the \emph{Geometry Expert}. 
This design provides a stable and explicit estimate of future geometry, which complements the implicit latent modeling in $\mathbf{z}_{t+1:t+H}$ and serves as guidance for geometry-aware action modulation.

\begin{figure}[t]
    \centering
    \includegraphics[width=\linewidth]{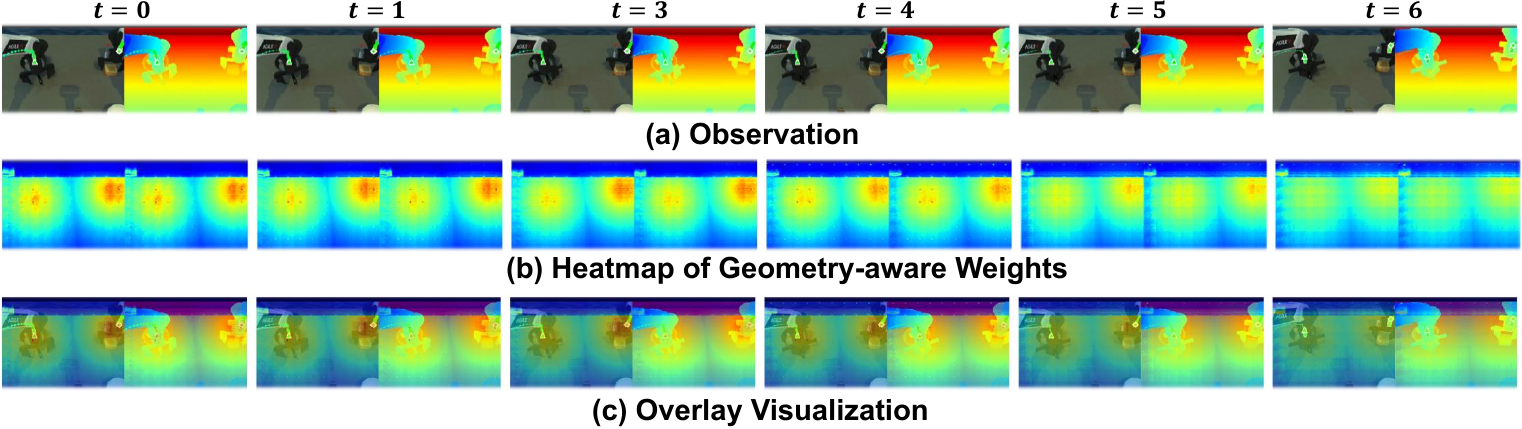}
    \caption{
    Geometry-aware weight construction and modulation in GASAM.
    Rows (a)--(c) show observations, geometry-aware weights, and their overlays across selected temporal steps.
    The constructed weights highlight geometrically action-relevant regions for action-attention modulation.
    }
    \label{fig:gasam_weight_modulation}
\end{figure}

\paragraph{Geometry-aware weight construction and modulation.}
To make attention aware of metric spatial relations, GASAM measures how close each visual location is to the predicted end-effector in 3D space. Given the predicted geometry, GASAM first recovers 3D points $\hat{\mathbf{P}}_{t,j}$ from the predicted depth $\hat{\mathbf{D}}_t$ using camera parameters, and computes their distances to the predicted end-effector position:
\begin{equation}
d_{t,j}
=
\left\|\hat{\mathbf{p}}_{t}-\hat{\mathbf{P}}_{t,j}\right\|_2 .
\end{equation}
The distances are converted into geometry-aware weights and aligned to the video-token grid:
\begin{equation}
\mathbf{w}_{t+1:t+H}
=
\mathcal{T}
\left(
\rho\left(\{d_{t,j}\}_{t,j}\right)
\right),
\end{equation}
where $\rho(\cdot)$ denotes a monotonically decreasing distance-to-weight mapping, and $\mathcal{T}(\cdot)$ samples or aggregates the resulting pixel-level weights onto the video-token grid.
As shown in Fig.~\ref{fig:gasam_weight_modulation}, the constructed geometry-aware weights highlight spatial regions that are geometrically relevant to action execution and provide the modulation signal for the action attention branch.

GASAM then uses $\mathbf{w}_{t+1:t+H}$ to modulate action-to-video attention. Let $\mathbf{Q}^{a}$ denote action queries, and let $\mathbf{K}^{v}$ and $\mathbf{V}^{v}$ denote video keys and values. The geometry-modulated action-to-video attention is:
\begin{equation}
\operatorname{Attn}^{a\leftarrow v}_{\mathrm{GASAM}}
=
\operatorname{Softmax}
\!\left(
\frac{\mathbf{Q}^{a}(\mathbf{K}^{v})^{\top}}{\sqrt{d}}
+
\lambda \log(\mathbf{w}_{t+1:t+H}+\epsilon)
\right)
\mathbf{V}^{v}.
\end{equation}
Here, $\lambda$ controls the modulation strength and $\epsilon$ avoids numerical instability. Since $\log(\mathbf{w}_{t+1:t+H}+\epsilon)$ is added before the softmax, GASAM directly biases the attention weights from action queries to visual tokens toward geometrically relevant regions. The modulation is applied only to the action branch, allowing geometry to guide action selection while preserving the original spatial-temporal modeling and visual-language understanding.

\subsection{Training and data}
\label{section3.4}


\paragraph{Initialization and diffusion objective.}

We initialize the \emph{ST World Model} from Wan~\cite{videodiffusion,latentdiffusion,dit,wan2025} and the \emph{Understanding Expert} from Qwen-VL~\cite{qwenvl}. The \emph{Action Expert} follows the same Transformer architecture as the \emph{ST World Model} but uses action-specific parameters, while the \emph{Geometry Expert} is implemented as a diffusion-aligned Transformer operating on the same video and action tokens to predict future geometry. We jointly model spatial-temporal observations $\mathbf{x}$ and actions $\mathbf{a}$ via diffusion by predicting velocity fields~\cite{ddpm,score_sde,flowmatching,diffusionpolicy}:
\begin{equation}
\mathcal{L}_{\text{obs}} =
\mathbb{E}\!\left[\|\mathbf{v}_{\theta}^{o}-(\boldsymbol{\epsilon}_{o}-\mathbf{x}_{t+1:t+H})\|^2\right], \quad
\mathcal{L}_{\text{action}} =
\mathbb{E}\!\left[\|\mathbf{v}_{\theta}^{a}-(\boldsymbol{\epsilon}_{a}-\mathbf{a}_{t+1:t+H})\|^2\right],
\end{equation}
where $\boldsymbol{\epsilon}_{o}$ and $\boldsymbol{\epsilon}_{a}$ denote Gaussian noise, and $\mathbf{v}_{\theta}^{o}$ and $\mathbf{v}_{\theta}^{a}$ are the predicted velocity fields. Different diffusion time steps ($\tau_v, \tau_a$) are used for observations and actions to account for modality differences.

The overall loss is:
\begin{equation}
\mathcal{L}_{\text{diff}}=\lambda_o \mathcal{L}_{\text{obs}}+\lambda_a \mathcal{L}_{\text{action}}.
\end{equation}

\paragraph{Geometry supervision.}
The \emph{Geometry Expert} is supervised on depth, end-effector position, and geometry-aware weights:
\begin{equation}
\mathcal{L}_{\text{geo}}=\lambda_d \mathcal{L}_{\text{depth}}+\lambda_p \mathcal{L}_{\text{pose}}+\lambda_w \mathcal{L}_{\text{weight}},
\end{equation}
where $\mathcal{L}_{\text{depth}}$ supervises depth prediction, $\mathcal{L}_{\text{pose}}$ constrains the 3D end-effector position, and $\mathcal{L}_{\text{weight}}$ enforces consistency between predicted modulation weights and geometry-derived targets.

The target weights are constructed from ground-truth spatial distances:
\begin{equation}
w^{*}_{t,j} = \varphi\!\left(\| \mathbf{p}_{t} - \mathbf{P}_{t,j} \|_2 \right), \quad
\mathcal{L}_{\text{weight}} =
\left\|
\mathbf{w}_{t+1:t+H} - \mathbf{w}^{*}_{t+1:t+H}
\right\|_2^2,
\end{equation}
where $w^{*}_{t,j}$ is derived from the distance between the end-effector and scene points, providing direct supervision for geometry-aware attention modulation.

The \emph{Geometry Expert} is first trained independently and then jointly fine-tuned, influencing action generation through GASAM rather than directly modifying the diffusion objective.

\paragraph{Training pipeline.}
We adopt a three-stage training strategy:

\textbf{Stage 1 (spatial-temporal pretraining):} train the \emph{ST World Model} and \emph{Understanding Expert} on large-scale video and multi-modal data to learn spatial-temporal dynamics and semantic representations. 

\textbf{Stage 2 (action and geometry learning):} introduce the \emph{Geometry Expert} and \emph{Action Expert} for joint training under the diffusion objective, enabling mapping from spatial-temporal representations to actions with geometric constraints. 

\textbf{Stage 3 (joint finetuning):} perform end-to-end finetuning with GASAM, allowing all modules to be optimized jointly for improved spatial-temporal consistency and action generation.

\paragraph{Dataset.}

\begin{table}[t] 
\caption{Hierarchical data organization for STARRY.}
\label{tab:pyramid}
\centering
\small
\begin{tabular}{lccccc}
\toprule
Level & Data Type & Language & Video & Geometry & Action \\
\midrule
L1 & Web-scale video & \checkmark & \checkmark &  &  \\
L2 & Egocentric video & \checkmark & \checkmark &  &  \\
L3 & Synthetic / simulation & \checkmark & \checkmark & \checkmark &  \\
L4 & Interaction data & \checkmark & \checkmark & \checkmark & \checkmark \\
L5 & Multi-robot trajectories & \checkmark & \checkmark & \checkmark & \checkmark \\
L6 & Target-robot data & \checkmark & \checkmark & \checkmark & \checkmark \\
\bottomrule
\end{tabular}
\end{table}

As shown in Table~\ref{tab:pyramid}, we organize data hierarchically to progressively introduce semantic, temporal, geometric, and action supervision. L1--L2 use large-scale web and egocentric videos, such as Ego4D~\cite{ego4d} and Ego-Dex~\cite{egodex}, to learn general visual features and temporal dynamics. L3--L4 introduce geometry-enriched data with depth and multi-view observations, such as EmbodiedMAE~\cite{embodiedmae}, for explicit spatial structure modeling. L5--L6 further incorporate real robot datasets~\cite{robomimic,droid,bridgedata,openx}, such as DROID and BridgeData V2, for action learning, with L6 finetuned on target-robot data for task-specific adaptation.

\section{Experiments}
We evaluate STARRY in simulation and real-world settings. Specifically, we test on RoboTwin 2.0 under Clean and Randomized settings (Sec.~\ref{sec:sim_eval}), examine physical execution in real-world experiments (Sec.~\ref{sec:real_eval}), and conduct ablations to analyze the effects of spatial-temporal prediction and GASAM (Sec.~\ref{sec:ablation}).

\subsection{Evaluation in simulation environment}
\label{sec:sim_eval}

\begin{table*}[t]
\centering
\small
\setlength{\tabcolsep}{4.6pt}
\caption{Selected RoboTwin 2.0 results under Clean and Randomized settings.}
\resizebox{\textwidth}{!}{
\begin{tabular}{lcccccccccc}
\toprule
\multirow{2}{*}{Task}
& \multicolumn{2}{c}{$\pi_{0.5}$}
& \multicolumn{2}{c}{X-VLA}
& \multicolumn{2}{c}{Motus}
& \multicolumn{2}{c}{LingBot-VA}
& \multicolumn{2}{c}{Ours} \\
& Clean & Rand. & Clean & Rand. & Clean & Rand. & Clean & Rand. & Clean & Rand. \\
\midrule
Click Alarmclock    & 97.5\% & 91\% & 99\% & 99\% & \textbf{100\%} & \textbf{100\%} & 99\% & \textbf{100\%} & \textbf{100\%} & \textbf{100\%} \\
Handover Mic        & 63\% & 57.5\% & 0\%  & 0\%  & 78\% & 63\% & 94\% & 96\% & \textbf{100\%} & \textbf{99\%} \\
Hanging Mug         & 10.5\% & 10\% & 23\% & 27\% & 38\% & 38\% & 40\% & 28\% & \textbf{69\%}  & \textbf{72\%} \\
Move Can Pot        & 40\% & 41\% & 89\% & 86\% & 34\% & 74\% & 94\% & 97\% & \textbf{100\%} & \textbf{98\%} \\
Move Pillbottle Pad & 58.5\% & 45\% & 73\% & 71\% & 93\% & 96\% & 99\% & 99\% & \textbf{100\%} & \textbf{100\%} \\
Pick Diverse Bottles& 43\% & 37\% & 58\% & 36\% & 90\% & 91\% & 89\% & 82\% & \textbf{98\%}  & \textbf{96\%} \\
Place Bread Basket  & 62.5\% & 60\% & 81\% & 71\% & 91\% & 94\% & 97\% & 95\% & \textbf{100\%} & \textbf{99\%} \\
\multicolumn{11}{c}{\textit{...... (50 tasks)}} \\
Place Can Basket    & 40.5\% & 43.5\% & 49\% & 52\% & 81\% & 76\% & 81\% & 84\% & \textbf{89\%}  & \textbf{88\%} \\
Place Shoe          & 74.5\% & 77\% & 96\% & 95\% & 99\% & 97\% & 98\% & 98\% & \textbf{100\%} & \textbf{99\%} \\
Press Stapler       & 83.5\% & 76.5\% & 92\% & 98\% & 93\% & 98\% & 85\% & 82\% & \textbf{100\%} & \textbf{100\%} \\
Put Bottles Dustbin & 48\% & 44\% & 74\% & 77\% & 81\% & 79\% & 87\% & 91\% & \textbf{96\%}  & \textbf{93\%} \\
Stack Blocks Three  & 53\% & 46\% & 6\%  & 10\% & 91\% & 95\% & \textbf{99\%} & 98\% & 97\% & \textbf{100\%} \\
Turn Switch         & 33.5\% & 30\% & 40\% & 61\% & 84\% & 78\% & 44\% & 45\% & \textbf{85\%}  & \textbf{89\%} \\
\midrule
Average             & 62.86\% & 60.30\% & 72.80\% & 72.84\% & 88.66\% & 87.02\% & 92.93\% & 91.55\% & \textbf{93.82\%} & \textbf{93.30\%} \\
\bottomrule
\end{tabular}
}
\label{tab:robotwin_main}
\end{table*}

\paragraph{RoboTwin 2.0}


We first evaluate STARRY on the RoboTwin 2.0 benchmark, which comprises 50 bimanual manipulation tasks with structured domain randomization~\cite{chen2025robotwin}. In the main paper, we compare against $\pi_{0.5}$~\cite{black2025pi05}, X-VLA~\cite{zheng2025xvla}, Motus~\cite{bi2025motus}, and LingBot-VA~\cite{li2026lingbotva}. For each task, we use 50 demonstrations under the Clean setting and 500 demonstrations under the Randomized setting. Demonstrations from all 50 tasks are pooled together and jointly optimized, rather than training each task independently. All models are trained with a batch size of 256 for 40k steps. To save space, additional baselines, including GO-1~\cite{bu2025agibotworld} and $\pi_0$~\cite{black2025pi0}, together with full per-task results on all 50 tasks, are provided in Appendix~\ref{app:complete_robotwin_results}. 

Overall, STARRY achieves the best average success rates under both Clean and Randomized settings, reaching 93.82\% and 93.30\%, respectively, surpassing LingBot-VA (92.93\% / 91.55\%) and Motus (88.66\% / 87.02\%). Notably, these average gains underestimate the advantage on challenging tasks, since many easier tasks are close to saturation for multiple methods. On more discriminative tasks, the improvements of STARRY align closely with our design. For example, on \textit{Handover Mic}, which requires precise temporal coordination in bimanual handover, STARRY achieves 100\% / 99\%, significantly outperforming Motus (78\% / 63\%), suggesting that spatial-temporal world modeling better supports anticipation of future interaction states. On tasks that rely on local geometric structure, contact location, and fine-grained alignment, such as \textit{Hanging Mug}, \textit{Turn Switch}, and \textit{Press Stapler}, STARRY obtains 69\% / 72\%, 85\% / 89\%, and 100\% / 100\%, respectively. In particular, on \textit{Hanging Mug}, STARRY improves the previous best result by more than 30 points, indicating that GASAM helps focus action generation on geometry-relevant regions. We also observe consistent gains on tasks involving object selection, container relations, and placement constraints, including \textit{Move Can Pot}, \textit{Pick Diverse Bottles}, \textit{Place Can Basket}, and \textit{Put Bottles Dustbin}. Taken together, these results show that the advantage of STARRY is not limited to average performance, but is most pronounced on tasks that truly require spatial-temporal reasoning and geometric modeling.


\begin{figure*}[t]
    \centering
    \includegraphics[width=\textwidth]{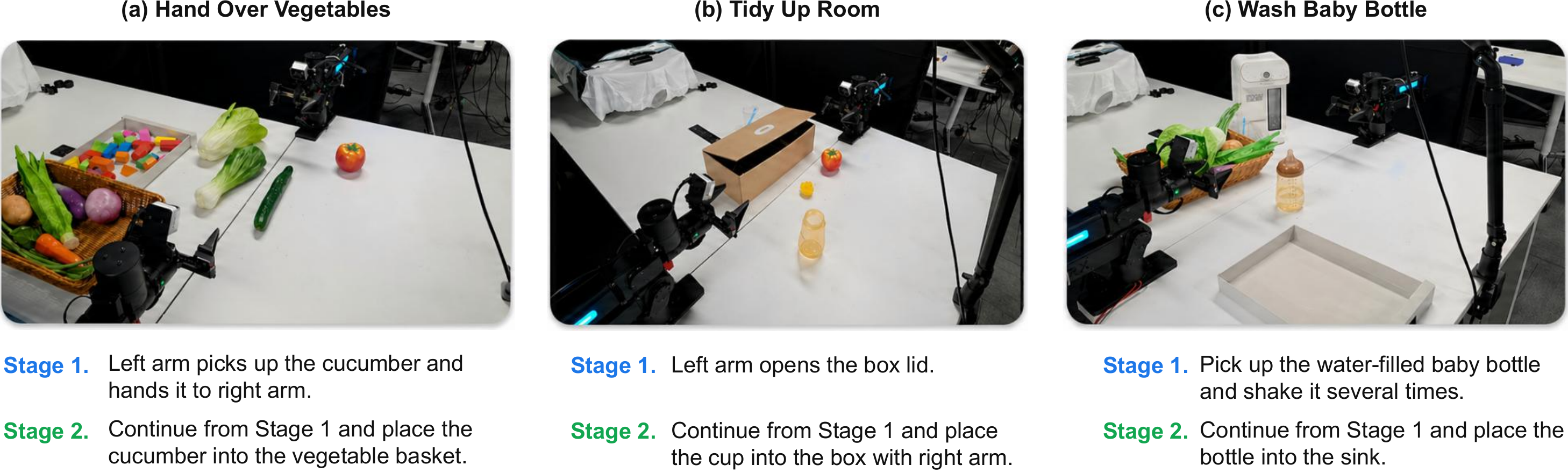}
    \caption{
    Real-world task settings. We evaluate three representative two-stage bimanual manipulation tasks: \textit{Hand Over Vegetables}, \textit{Tidy Up Room}, and \textit{Wash Baby Bottle}.
    }
    \label{fig:real_world_tasks}
\end{figure*}

\subsection{Real-world experiments}
\label{sec:real_eval}

To further evaluate the effectiveness of STARRY in real robotic settings, we conduct real-world experiments on representative bimanual manipulation tasks and compare against $\pi_{0.5}$~\cite{black2025pi05}. Compared with simulation, real-world scenarios involve stronger perception noise, execution errors, and interaction uncertainty, and therefore provide a more direct measure of practical deployment performance. We consider three tasks, namely \textit{Hand Over Vegetables}, \textit{Tidy Up Room}, and \textit{Wash Baby Bottle}, as shown in Fig.~\ref{fig:real_world_tasks}. To better characterize policy capability, each task is evaluated in two stages: Stage 1 focuses on the key subskill, while Stage 2 requires completing the full multi-step task built upon Stage 1. All real-world experiments are conducted on the ARX R5 bimanual robot platform from ARX. For each task, we collect 50 real-robot demonstrations for training and evaluate each method with 20 rollouts, reporting success rate as the metric.

\begin{table}[t]
\centering
\small
\setlength{\tabcolsep}{5pt}
\caption{Success rates (\%) in real-world experiments.}
\begin{tabular}{lcccccc}
\toprule
\multirow{2}{*}{Method}
& \multicolumn{2}{c}{Hand Over Vegetables}
& \multicolumn{2}{c}{Tidy Up Room}
& \multicolumn{2}{c}{Wash Baby Bottle} \\
& Stage 1 & Stage 2 & Stage 1 & Stage 2 & Stage 1 & Stage 2 \\
\midrule
$\pi_{0.5}$ & 60\% & 40\% & 55\% & 35\% & 40\% & 25\% \\
Ours        & \textbf{85\%} & \textbf{70\%} & \textbf{75\%} & \textbf{65\%} & \textbf{70\%} & \textbf{60\%} \\
\bottomrule
\end{tabular}
\label{tab:real_world_results}
\end{table}

\begin{figure*}[t]
    \centering
    \includegraphics[width=\textwidth]{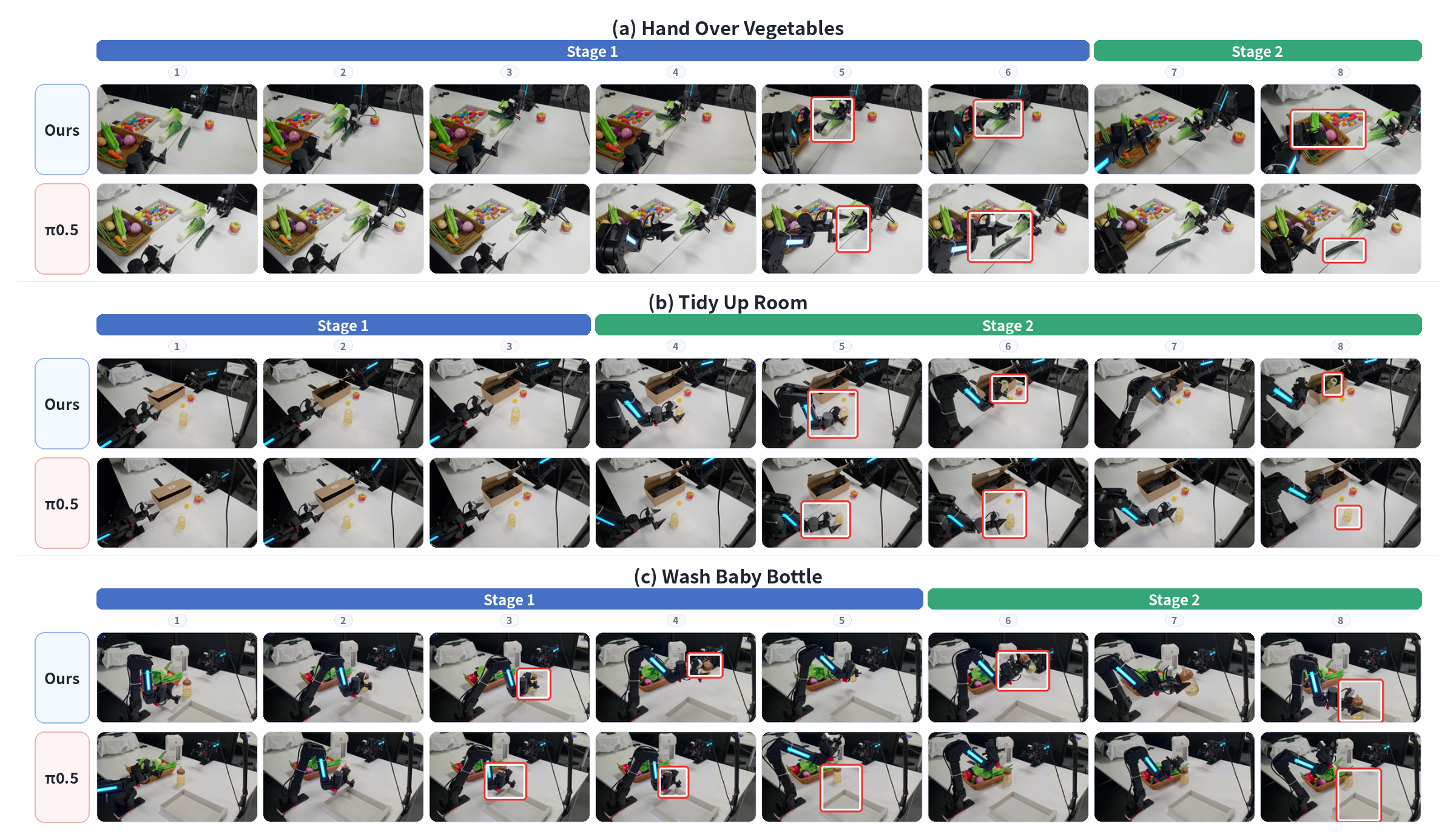}
    \caption{Qualitative comparison in real-world experiments. STARRY and $\pi_{0.5}$ are compared on three bimanual tasks. Red boxes highlight key regions related to success or failure.}
    \label{fig:real_world_compare}
\end{figure*}

Table~\ref{tab:real_world_results} reports the real-world success rates. STARRY achieves 70.8\% average success over the six evaluations, compared with 42.5\% for $\pi_{0.5}$, consistently outperforming it across all tasks and stages. 
Notably, the gain becomes larger in the more challenging Stage 2 setting, where STARRY improves success by 31.7 percentage points, suggesting that its advantage is not limited to stabilizing short-horizon actions but also comes from stronger spatial-temporal modeling for multi-step execution.
The qualitative comparisons in Fig.~\ref{fig:real_world_compare} further support this observation. In \textit{Hand Over Vegetables} (Fig.~\ref{fig:real_world_compare}(a)), $\pi_{0.5}$ often fails to establish a proper spatial-temporal configuration for bimanual handover, leading to grasping or transfer errors, whereas STARRY completes the sequence more reliably. In \textit{Tidy Up Room} (Fig.~\ref{fig:real_world_compare}(b)), $\pi_{0.5}$ may open the box but fail to align the object with the box opening during placement, while STARRY better preserves the target-container spatial relationship. In \textit{Wash Baby Bottle} (Fig.~\ref{fig:real_world_compare}(c)), the irregular bottle geometry makes grasping, shaking, and placement sensitive to local contact and temporal coordination; STARRY successfully maintains a stable grasp, executes the shaking motion, and completes the final placement. These results are consistent with our design: jointly denoised spatial-temporal latents provide coherent future interaction cues, while the \emph{Geometry Expert} and GASAM inject action-aligned geometric guidance into critical local regions, improving spatial-temporal coordination, object alignment, and multi-step execution in real-world bimanual manipulation.

\begin{table}[t]
\centering
\small
\setlength{\tabcolsep}{4.2pt}
\caption{
Ablation on predictive representation and geometry-aware modulation on RoboTwin 2.0.
Values in parentheses denote absolute gains from enabling GASAM under the same predictive representation.
\textbf{Act.}: Action-Only Denoising.
\textbf{App.}: Appearance-Only Prediction.
\textbf{ST}: Full Spatial-Temporal Modeling.
}
\label{tab:ablation_overall}
\begin{tabular}{lcccccc}
\toprule
\multirow{2}{*}{GASAM}
& \multicolumn{3}{c}{Randomized}
& \multicolumn{3}{c}{Clean} \\
\cmidrule(lr){2-4}\cmidrule(lr){5-7}
& Act. & App. & ST
& Act. & App. & ST \\
\midrule
Disabled
& 64.96\% & 85.80\% & 88.82\%
& 63.42\% & 86.64\% & 90.40\% \\
Enabled
& 75.88\% {\scriptsize (+10.92)}
& 86.96\% {\scriptsize (+1.16)}
& \textbf{93.30\%} {\scriptsize (+4.48)}
& 72.50\% {\scriptsize (+9.08)}
& 87.86\% {\scriptsize (+1.22)}
& \textbf{93.82\%} {\scriptsize (+3.42)} \\
\midrule
$\Delta_{\mathrm{ST}-\mathrm{Act}}$
& \multicolumn{3}{c}{+23.86 w/o GASAM;\quad +17.42 w/ GASAM}
& \multicolumn{3}{c}{+26.98 w/o GASAM;\quad +21.32 w/ GASAM} \\
\bottomrule
\end{tabular}
\end{table}

\subsection{Ablation study}
\label{sec:ablation}

Table~\ref{tab:ablation_overall} studies the effects of predictive representation and geometry-aware modulation.
Without GASAM, moving from action-only denoising to appearance-only prediction significantly improves performance from 64.96\% to 85.80\% on Randomized and from 63.42\% to 86.64\% on Clean, showing that future appearance prediction provides useful temporal context beyond reactive action generation.
Further introducing full spatial-temporal modeling improves the results to 88.82\% and 90.40\%, respectively.
This indicates that appearance prediction alone is insufficient for manipulation: explicitly modeling trajectory evolution and spatial geometry provides additional constraints for contact, handover, and placement, which are critical in spatially demanding manipulation tasks.

Enabling GASAM consistently improves all representation variants.
The gain is especially large for action-only denoising (+10.92\% on Randomized and +9.08\% on Clean), suggesting that geometry-aware weights provide an effective spatial prior even without full future prediction.
Combined with ST modeling, GASAM further improves performance from 88.82\% to 93.30\% on Randomized and from 90.40\% to 93.82\% on Clean, achieving the best overall results.
The smaller gain on appearance-only prediction may be because appearance prediction already captures coarse object saliency but lacks precise geometric grounding.
Task-level results in Appendix~\ref{app:complete_robotwin_results} further show larger improvements on spatially sensitive tasks such as handover and placement, supporting that STARRY benefits from both spatial-temporal prediction and geometry-aware action-attention modulation.

\section{Conclusion}

We presented \textbf{STARRY}, a world-model-enhanced action-generation policy that aligns spatial-temporal prediction with action generation. STARRY jointly denoises future spatial-temporal latents and action sequences, and uses GASAM to inject predicted geometry into the action attention branch. On RoboTwin 2.0, STARRY achieves \textbf{93.82\%} / \textbf{93.30\%} average success under Clean and Randomized settings. In real-world experiments, it improves average success from \textbf{42.5\%} to \textbf{70.8\%} compared with $\pi_{0.5}$, demonstrating improved execution reliability in physical manipulation. These results demonstrate the effectiveness of action-centric spatial-temporal world modeling for spatial-temporally demanding robotic action generation.

\begin{ack}
We thank Yuzhuo Miao for his valuable support in dataset collection and processing. We also thank Yunlong Guo, Zhaolong Shen, Shijie Lian, and Xiaopeng Lin for their assistance with data collection.
\end{ack}


\bibliographystyle{plain}
\bibliography{refs.bib}


\newpage
\appendix

\section{Additional details and pseudocode}
\label{app:technical_details}

\subsection{Hyperparameters}
\label{app:implementation_details}

We summarize the key numerical hyperparameters used in STARRY in Table~\ref{tab:implementation_details}. The model uses branch-specific diffusion timesteps for video and action denoising, while the Geometry Expert predicts depth and multicamera XYZ end-effector positions to provide GASAM-based action modulation.

\begin{table}[p]
\centering
\small
\setlength{\tabcolsep}{6pt}
\caption{Key implementation hyperparameters of STARRY.}
\label{tab:implementation_details}
\begin{tabular}{llc}
\toprule
Component & Hyperparameter & Value \\
\midrule
\multirow{6}{*}{Policy}
& State / action dim. & $14$ / $14$ \\
& Input resolution & $384 \times 640$ \\
& Predicted video frames & $8$ \\
& Action chunk size & $16$ \\
& Inference denoising steps & $10$ \\
& Observation / action loss weight & $1.0$ / $2.0$ \\
\midrule
\multirow{6}{*}{ST World Model}
& Hidden size / FFN dim. & $3072$ / $14336$ \\
& Layers / heads & $30$ / $24$ \\
& Input / output dim. & $48$ / $48$ \\
& Frequency embedding dim. & $256$ \\
& Text length & $512$ \\
& Layer norm epsilon & $1\times10^{-6}$ \\
\midrule
\multirow{3}{*}{Action Expert}
& Hidden size & $1024$ \\
& Layers / heads & $30$ / $16$ \\
& FFN multiplier / norm epsilon & $4$ / $1\times10^{-6}$ \\
\midrule
\multirow{2}{*}{Understanding Expert}
& Hidden size & $512$ \\
& FFN multiplier / norm epsilon & $4$ / $1\times10^{-5}$ \\
\midrule
\multirow{4}{*}{Geometry Expert}
& Video / action token dim. & $3072$ / $1024$ \\
& Hidden size / camera dim. & $512$ / $256$ \\
& Number of cameras & $3$ \\
& Depth / pose / target loss weight & $1.0$ / $1.0$ / $1.0$ \\
\midrule
\multirow{4}{*}{Training}
& Optimizer & AdamW \\
& Training steps & $40$k \\
& Learning rate & $1$--$5\times10^{-5}$ \\
& Weight decay & $0.01$ \\
\bottomrule
\end{tabular}
\end{table}

\subsection{Compute resources}
\label{app:resources}
For fine-tuning and evaluation, we use GPU workers equipped with 8 NVIDIA Tesla A100-80G GPUs. Each worker provides 112 vCPUs, 1960 GiB memory, 80 Gbit/s network bandwidth, and 32 Gbit/s disk bandwidth. A typical fine-tuning run with batch size 256 for 40k steps takes approximately one week on this configuration. Large-scale pretraining uses the same type of GPU worker but scales to more nodes depending on the training stage and data scale. The main reported experiments, including RoboTwin 2.0 evaluation and real-world policy evaluation, are conducted with this A100-based configuration.

\subsection{Pseudocode of STARRY}
\label{app:pseudocode}

Algorithm~\ref{alg:starry_training} summarizes the training procedure of STARRY. For each training batch, we first construct spatial-temporal inputs from multi-view RGB-D observations and projected end-effector trajectories, and encode them into video tokens. The \emph{Understanding Expert} provides semantic grounding tokens, while the \emph{ST World Model} and \emph{Action Expert} jointly denoise future spatial-temporal latents and action sequences under branch-specific diffusion timesteps. In parallel, the \emph{Geometry Expert} predicts future depth and end-effector positions, from which GASAM computes token-aligned geometry-aware weights. These weights are injected only into the action attention branch, so geometric priors guide action generation without perturbing video modeling or visual-language understanding.

\begin{algorithm}[p]
\caption{STARRY training with spatial-temporal denoising and GASAM}
\label{alg:starry_training}
\begin{algorithmic}[1]
\Require Training batch $\mathcal{B}=\{(\mathbf{o}_{t_0:t+H},\mathbf{a}_{t+1:t+H})\}$
\Require ST World Model $f_{\theta}^{\mathrm{ST}}$, Action Expert $f_{\theta}^{a}$, Geometry Expert $g_{\phi}$, Understanding Expert $h_{\psi}$
\Ensure Optimized parameters $\theta,\phi,\psi$

\For{each training iteration}
    \State Sample a mini-batch from $\mathcal{B}$

    \State Project end-effector trajectories into camera views as $\mathbf{u}_{\tau}^{c,m}=\Pi(K_t^c,T_t^c,\mathbf{e}_{\tau}^{m})$, and construct spatial-temporal inputs:
    \Statex \hspace{\algorithmicindent}
    \[
    \mathbf{x}_t =
    \Phi(\{\mathbf{I}_t^{c}\}_{c},
    \{\mathbf{D}_t^{c}\}_{c},
    \{\mathbf{u}_{\tau}^{c,m}\}_{\tau\le t,c,m})
    \]

    \State Encode $\mathbf{X}_{t_0:t}$ into video tokens $\mathbf{v}_{t_0:t}$ and obtain understanding tokens $\mathbf{s}=h_{\psi}(\mathbf{o}_t)$

    \State Sample branch-specific diffusion timesteps $(\tau_v,\tau_a)$ and corrupt future spatial-temporal targets and action targets

    \State Predict future geometry with the Geometry Expert:
    \Statex \hspace{\algorithmicindent}
    \[
    (\hat{\mathbf{D}}_{t+1:t+H},\hat{\mathbf{p}}_{t+1:t+H})
    =
    g_{\phi}(\mathbf{v}_{t_0:t},\mathbf{a}_{\le t},\mathbf{o}_t,\tau_v,\tau_a)
    \]

    \State Compute geometry-aware weights $\mathbf{w}_{t+1:t+H}$ from predicted 3D distances between $\hat{\mathbf{p}}_t$ and scene points $\hat{\mathbf{P}}_{t,j}$

    \State Apply GASAM only to action-to-video attention:
    \Statex \hspace{\algorithmicindent}
    \[
    \operatorname{Attn}^{a\leftarrow v}_{\mathrm{GASAM}}
    =
    \operatorname{Softmax}
    \!\left(
    \frac{\mathbf{Q}^{a}(\mathbf{K}^{v})^{\top}}{\sqrt{d}}
    +
    \lambda\log(\mathbf{w}_{t+1:t+H}+\epsilon)
    \right)
    \mathbf{V}^{v}
    \]

    \State Jointly denoise future spatial-temporal latents and actions with GASAM-modulated action attention:
    \Statex \hspace{\algorithmicindent}
    \[
    (\hat{\mathbf{z}}_{t+1:t+H},\hat{\mathbf{a}}_{t+1:t+H})
    =
    (f_{\theta}^{\mathrm{ST}},f_{\theta}^{a})
    (\mathbf{v}_{t_0:t},\mathbf{s},\tau_v,\tau_a)
    \]

    \State Compute diffusion loss $\mathcal{L}_{\mathrm{diff}}=\lambda_o\mathcal{L}_{\mathrm{obs}}+\lambda_a\mathcal{L}_{\mathrm{action}}$
    \State Compute geometry loss $\mathcal{L}_{\mathrm{geo}}=\lambda_d\mathcal{L}_{\mathrm{depth}}+\lambda_p\mathcal{L}_{\mathrm{pose}}+\lambda_w\mathcal{L}_{\mathrm{weight}}$

    \State Update parameters using
    $\mathcal{L}=\mathcal{L}_{\mathrm{diff}}+\mathcal{L}_{\mathrm{geo}}$
\EndFor
\end{algorithmic}
\end{algorithm}

During inference, STARRY follows the same denoising structure but removes all supervision terms. Starting from noisy future latent and action variables, the model iteratively predicts future geometry, constructs GASAM weights, and updates the latent-action pair through joint denoising. The final action sequence $\mathbf{a}_{t+1:t+H}$ is executed by the robot, while the predicted spatial-temporal latents and geometry serve as internal guidance for action generation.

\section{Additional experimental results}
\label{app:additional_results}

\subsection{Visualization of future spatial-temporal latents}
\label{app:future_latent_vis}

To further illustrate what is captured by the learned future spatial-temporal latents, we visualize decoded future predictions in both simulation and real-world settings. These visualizations are not used as additional supervision during evaluation, but serve to qualitatively examine whether the predicted latents preserve temporally coherent scene evolution, end-effector motion, and spatial geometry. As shown below, STARRY predicts not only future appearance changes, but also structured depth and geometry patterns that are closely aligned with manipulation-relevant regions.

\begin{figure*}[p]
\centering
\includegraphics[width=\textwidth]{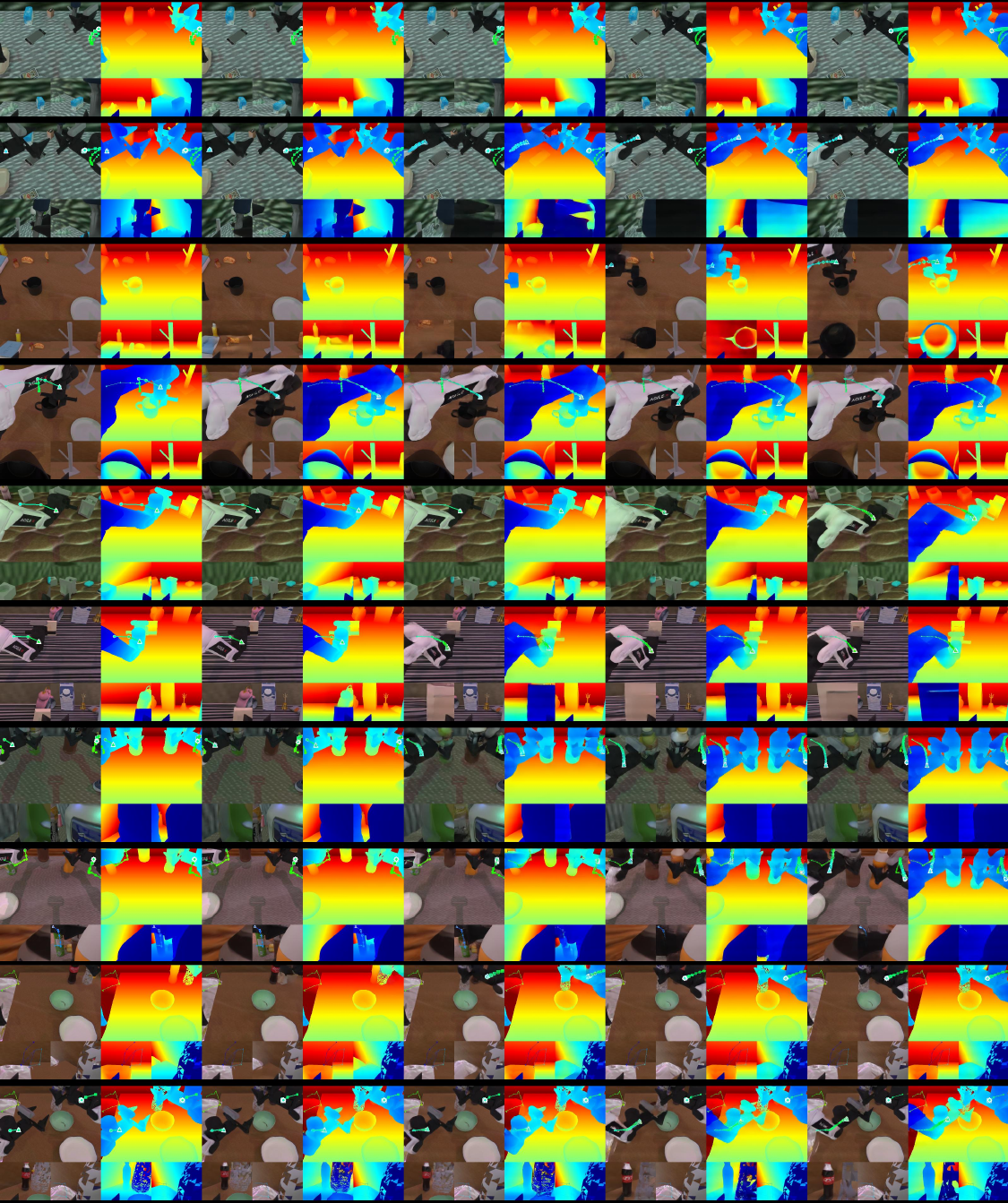}
\caption{
Visualization of future spatial-temporal latents on RoboTwin 2.0. 
Each row corresponds to a manipulation rollout, and columns show consecutive future predictions decoded from the spatial-temporal latent representation. 
The predicted latents preserve object layouts, end-effector motion trends, and depth-related spatial structure across future steps, indicating that the ST World Model captures more than appearance-level temporal consistency.
}
\label{fig:app_robotwin_future_latents}
\end{figure*}

Fig.~\ref{fig:app_robotwin_future_latents} shows representative visualizations on RoboTwin 2.0. The decoded predictions maintain coherent future scene evolution across diverse simulated tasks, including changes in object positions, manipulator configurations, and depth structure. This supports our claim that the ST World Model learns a structured spatial-temporal representation rather than only predicting visually plausible RGB frames.

\begin{figure*}[t]
\centering
\includegraphics[width=\textwidth]{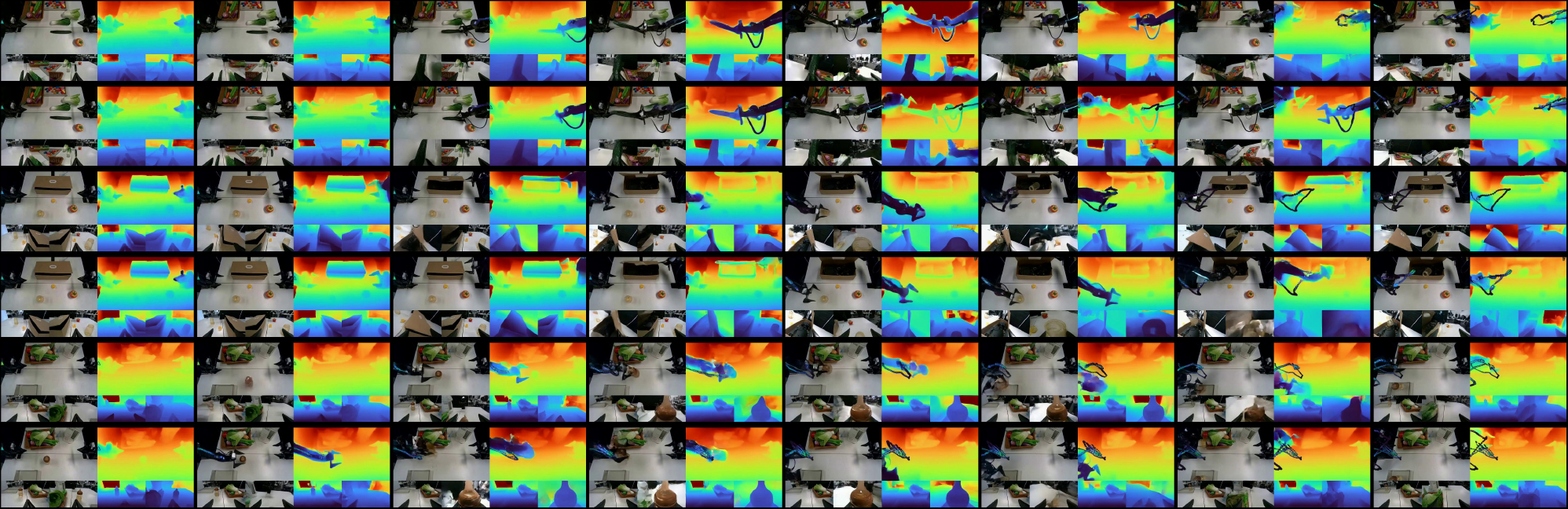}
\caption{
Visualization of future spatial-temporal latents in real-world manipulation. 
The decoded predictions show temporally consistent RGB-D structure under physical robot execution, including future end-effector motion and task-relevant spatial changes.
}
\label{fig:app_real_future_latents}
\end{figure*}

Fig.~\ref{fig:app_real_future_latents} further visualizes future spatial-temporal latents in real-world tasks. Despite real-world visual noise, lighting variation, and cluttered backgrounds, the predictions remain temporally coherent and preserve task-relevant spatial layouts. These results indicate that the learned latent representation provides useful predictive guidance for physical manipulation, complementing the quantitative real-world results reported in the main paper.

\subsection{Complete RoboTwin 2.0 results}
\label{app:complete_robotwin_results}

Table~\ref{tab:robotwin_merged} reports the complete task-level results on all 50 RoboTwin 2.0 tasks under Clean and Randomized settings. STARRY achieves the best average success rates, reaching \textbf{93.82\%} / \textbf{93.30\%}.

Table~\ref{tab:ablation_detailed} provides the complete task-level ablation results. Full spatial-temporal modeling with GASAM achieves the strongest performance, improving from 88.82\% to 93.30\% under Randomized and from 90.40\% to 93.82\% under Clean compared with disabling GASAM. These results further validate the complementary effects of spatial-temporal modeling and geometry-aware action modulation.

\begin{table*}[p]
\centering
\scriptsize
\small
\setlength{\tabcolsep}{3.2pt}
\caption{Complete task-level results on RoboTwin 2.0 under Clean and Randomized settings.
We report success rates over all 50 tasks for each method.
Bold numbers indicate the best result for each task and setting; ties are all bolded.}
\resizebox{\textwidth}{!}{
\begin{tabular}{lcccccccccccccc}
\toprule
\multirow{2}{*}{Simulation Task} & \multicolumn{2}{c}{GO-1} & \multicolumn{2}{c}{$\pi_0$} & \multicolumn{2}{c}{$\pi_{0.5}$} & \multicolumn{2}{c}{X-VLA} & \multicolumn{2}{c}{Motus} & \multicolumn{2}{c}{LingBot-VA} & \multicolumn{2}{c}{Ours} \\
& Clean & Rand. & Clean & Rand. & Clean & Rand. & Clean & Rand. & Clean & Rand. & Clean & Rand. & Clean & Rand. \\
\midrule
Adjust Bottle & 49\% & 62\% & 99\% & 95\% & 89.5\% & 91\% & \textbf{100\%} & \textbf{99\%} & 89\% & 93\% & 90\% & 94\% & 98\% & 97\% \\
Beat Block Hammer & 6\% & 10\% & 79\% & 84\% & 79.5\% & 71.5\% & 92\% & 88\% & 95\% & 88\% & \textbf{96\%} & \textbf{98\%} & 92\% & 89\% \\
Blocks Ranking Rgb & 7\% & 3\% & 80\% & 63\% & 67.5\% & 60\% & 83\% & 83\% & 99\% & 97\% & 99\% & \textbf{98\%} & \textbf{100\%} & \textbf{98\%} \\
Blocks Ranking Size & 2\% & 2\% & 14\% & 5\% & 28.5\% & 20\% & 67\% & 74\% & 75\% & 63\% & \textbf{94\%} & \textbf{96\%} & 71\% & 74\% \\
Click Alarmclock & 95\% & 90\% & 77\% & 68\% & 97.5\% & 91\% & 99\% & 99\% & \textbf{100\%} & \textbf{100\%} & 99\% & \textbf{100\%} & \textbf{100\%} & \textbf{100\%} \\
Click Bell & 98\% & 95\% & 71\% & 48\% & 87\% & 71\% & \textbf{100\%} & \textbf{100\%} & \textbf{100\%} & \textbf{100\%} & \textbf{100\%} & \textbf{100\%} & \textbf{100\%} & \textbf{100\%} \\
Dump Bin Bigbin & 57\% & 45\% & 88\% & 83\% & 61\% & 69.5\% & 79\% & 77\% & \textbf{95\%} & 91\% & 89\% & 96\% & 93\% & \textbf{97\%} \\
Grab Roller & 99\% & 99\% & 98\% & 94\% & 95\% & 94.5\% & \textbf{100\%} & \textbf{100\%} & \textbf{100\%} & \textbf{100\%} & \textbf{100\%} & \textbf{100\%} & \textbf{100\%} & \textbf{100\%} \\
Handover Block & 9\% & 12\% & 47\% & 31\% & 42\% & 38\% & 73\% & 37\% & 86\% & 73\% & \textbf{99\%} & \textbf{78\%} & 56\% & 48\% \\
Handover Mic & 12\% & 8\% & 97\% & 97\% & 63\% & 57.5\% & 0\% & 0\% & 78\% & 63\% & 94\% & 96\% & \textbf{100\%} & \textbf{99\%} \\
Hanging Mug & 0\% & 0\% & 14\% & 11\% & 10.5\% & 10\% & 23\% & 27\% & 38\% & 38\% & 40\% & 28\% & \textbf{69\%} & \textbf{72\%} \\
Lift Pot & 92\% & 92\% & 80\% & 72\% & 48\% & 42.5\% & 99\% & \textbf{100\%} & 96\% & 99\% & \textbf{100\%} & 99\% & \textbf{100\%} & \textbf{100\%} \\
Move Can Pot & 16\% & 4\% & 68\% & 48\% & 40\% & 41\% & 89\% & 86\% & 34\% & 74\% & 94\% & 97\% & \textbf{100\%} & \textbf{98\%} \\
Move Pillbottle Pad & 9\% & 11\% & 67\% & 46\% & 58.5\% & 45\% & 73\% & 71\% & 93\% & 96\% & 99\% & 99\% & \textbf{100\%} & \textbf{100\%} \\
Move Playingcard Away & 37\% & 24\% & 74\% & 65\% & 77.5\% & 75.5\% & 93\% & 98\% & \textbf{100\%} & 96\% & \textbf{100\%} & 99\% & \textbf{100\%} & \textbf{100\%} \\
Move Stapler Pad & 3\% & 4\% & 41\% & 24\% & 36\% & 30\% & 78\% & 73\% & 83\% & 85\% & 91\% & 79\% & \textbf{93\%} & \textbf{93\%} \\
Open Laptop & 65\% & 60\% & 71\% & 81\% & 54.5\% & 65.5\% & 93\% & \textbf{100\%} & \textbf{95\%} & 91\% & 92\% & 94\% & 93\% & 91\% \\
Open Microwave & 12\% & 14\% & 4\% & 32\% & 34.5\% & 57\% & 79\% & 71\% & \textbf{95\%} & \textbf{91\%} & 82\% & 86\% & 58\% & 61\% \\
Pick Diverse Bottles & 61\% & 56\% & 69\% & 31\% & 43\% & 37\% & 58\% & 36\% & 90\% & 91\% & 89\% & 82\% & \textbf{98\%} & \textbf{96\%} \\
Pick Dual Bottles & 81\% & 74\% & 59\% & 37\% & 51.5\% & 34.5\% & 47\% & 36\% & 96\% & 90\% & \textbf{100\%} & 99\% & \textbf{100\%} & \textbf{100\%} \\
Place A2b Left & 33\% & 36\% & 43\% & 47\% & 74.5\% & 71\% & 48\% & 49\% & 85\% & 79\% & \textbf{97\%} & \textbf{93\%} & 96\% & 91\% \\
Place A2b Right & 31\% & 22\% & 39\% & 34\% & 74.5\% & 70.5\% & 36\% & 36\% & 90.5\% & 87\% & \textbf{97\%} & 95\% & 94\% & \textbf{96\%} \\
Place Bread Basket & 47\% & 52\% & 62\% & 46\% & 62.5\% & 60\% & 81\% & 71\% & 91\% & 94\% & 97\% & 95\% & \textbf{100\%} & \textbf{99\%} \\
Place Bread Skillet & 2\% & 1\% & 66\% & 49\% & 61.5\% & 56\% & 77\% & 67\% & 86\% & 83\% & 95\% & 90\% & \textbf{96\%} & \textbf{95\%} \\
Place Burger Fries & 88\% & 92\% & 81\% & 76\% & 80\% & 78.5\% & 94\% & 94\% & 98\% & 98\% & 97\% & 95\% & \textbf{99\%} & \textbf{99\%} \\
Place Can Basket & 29\% & 37\% & 55\% & 46\% & 40.5\% & 43.5\% & 49\% & 52\% & 81\% & 76\% & 81\% & 84\% & \textbf{89\%} & \textbf{88\%} \\
Place Cans Plasticbox & 68\% & 77\% & 63\% & 45\% & 67\% & 65.5\% & 97\% & 98\% & 98\% & 94\% & \textbf{100\%} & 99\% & \textbf{100\%} & \textbf{100\%} \\
Place Container Plate & 73\% & 70\% & 97\% & 92\% & 85\% & 86.5\% & 97\% & 95\% & 98\% & 99\% & 99\% & 97\% & \textbf{100\%} & \textbf{100\%} \\
Place Dual Shoes & 6\% & 10\% & 59\% & 51\% & 43.5\% & 41\% & 79\% & 88\% & 93\% & 87\% & 94\% & 89\% & \textbf{96\%} & \textbf{94\%} \\
Place Empty Cup & 44\% & 39\% & 91\% & 85\% & 87.5\% & 92.5\% & \textbf{100\%} & 98\% & 99\% & 98\% & \textbf{100\%} & \textbf{100\%} & \textbf{100\%} & \textbf{100\%} \\
Place Fan & 1\% & 0\% & 66\% & 71\% & 56\% & 60.5\% & 80\% & 75\% & 91\% & 87\% & \textbf{99\%} & \textbf{93\%} & 95\% & 92\% \\
Place Mouse Pad & 15\% & 10\% & 20\% & 20\% & 40.5\% & 32.5\% & 70\% & 70\% & 66\% & 68\% & 93\% & 96\% & \textbf{96\%} & \textbf{98\%} \\
Place Object Basket & 48\% & 49\% & 67\% & 70\% & 61.5\% & 56\% & 44\% & 39\% & 81\% & 87\% & \textbf{91\%} & 88\% & 89\% & \textbf{91\%} \\
Place Object Scale & 26\% & 27\% & 57\% & 52\% & 63\% & 64.5\% & 52\% & 74\% & 88\% & 85\% & \textbf{96\%} & \textbf{95\%} & 89\% & 88\% \\
Place Object Stand & 56\% & 63\% & 82\% & 68\% & 82.5\% & 75\% & 86\% & 88\% & 98\% & 97\% & 99\% & 96\% & \textbf{100\%} & \textbf{100\%} \\
Place Phone Stand & 30\% & 37\% & 49\% & 53\% & 65\% & 67\% & 88\% & 87\% & 87\% & 86\% & \textbf{97\%} & \textbf{97\%} & 91\% & \textbf{97\%} \\
Place Shoe & 15\% & 13\% & 76\% & 76\% & 74.5\% & 77\% & 96\% & 95\% & 99\% & 97\% & 98\% & 98\% & \textbf{100\%} & \textbf{99\%} \\
Press Stapler & 66\% & 51\% & 44\% & 37\% & 83.5\% & 76.5\% & 92\% & 98\% & 93\% & 98\% & 85\% & 82\% & \textbf{100\%} & \textbf{100\%} \\
Put Bottles Dustbin & 7\% & 4\% & 65\% & 56\% & 48\% & 44\% & 74\% & 77\% & 81\% & 79\% & 87\% & 91\% & \textbf{96\%} & \textbf{93\%} \\
Put Object Cabinet & 60\% & 43\% & 73\% & 60\% & 52\% & 47\% & 46\% & 48\% & \textbf{88\%} & 71\% & 85\% & \textbf{87\%} & 84\% & 77\% \\
Rotate Qrcode & 22\% & 9\% & 74\% & 70\% & 68\% & 71.5\% & 34\% & 33\% & 89\% & 73\% & \textbf{96\%} & \textbf{91\%} & 94\% & 84\% \\
Scan Object & 1\% & 2\% & 55\% & 42\% & 57\% & 51.5\% & 14\% & 36\% & 67\% & 66\% & \textbf{96\%} & 91\% & 93\% & \textbf{96\%} \\
Shake Bottle Horizontally & 97\% & 92\% & 98\% & 92\% & 97.5\% & 99.5\% & \textbf{100\%} & \textbf{100\%} & \textbf{100\%} & 98\% & \textbf{100\%} & 99\% & \textbf{100\%} & 99\% \\
Shake Bottle & 97\% & 93\% & 94\% & 91\% & 95\% & 98.5\% & 99\% & \textbf{100\%} & \textbf{100\%} & 97\% & \textbf{100\%} & 97\% & \textbf{100\%} & 97\% \\
Stack Blocks Three & 1\% & 1\% & 72\% & 52\% & 53\% & 46\% & 6\% & 10\% & 91\% & 95\% & \textbf{99\%} & 98\% & 97\% & \textbf{100\%} \\
Stack Blocks Two & 12\% & 22\% & 93\% & 79\% & 72.5\% & 78\% & 92\% & 87\% & \textbf{100\%} & 98\% & \textbf{100\%} & 98\% & \textbf{100\%} & \textbf{100\%} \\
Stack Bowls Three & 4\% & 7\% & 77\% & 75\% & 55\% & 53\% & 76\% & 86\% & 79\% & 87\% & 86\% & 83\% & \textbf{91\%} & \textbf{92\%} \\
Stack Bowls Two & 51\% & 45\% & 94\% & 95\% & 86.5\% & 81\% & 96\% & 93\% & 98\% & 98\% & 94\% & 98\% & \textbf{100\%} & \textbf{99\%} \\
Stamp Seal & 19\% & 13\% & 46\% & 33\% & 57.5\% & 39\% & 76\% & 82\% & 93\% & 92\% & 96\% & 97\% & \textbf{100\%} & \textbf{99\%} \\
Turn Switch & 34\% & 30\% & 41\% & 42\% & 33.5\% & 30\% & 40\% & 61\% & 84\% & 78\% & 44\% & 45\% & \textbf{85\%} & \textbf{89\%} \\
Average (\%) & 37.8\% & 36.24\% & 65.92\% & 58.4\% & 62.86\% & 60.3\% & 72.8\% & 72.84\% & 88.66\% & 87.02\% & 92.93\% & 91.55\% & \textbf{93.82\%} & \textbf{93.3\%} \\
\bottomrule
\end{tabular}
}
\label{tab:robotwin_merged}
\end{table*}

\begin{table*}[t]
\centering
\scriptsize
\setlength{\tabcolsep}{2.8pt}
\caption{
Task-level ablation on RoboTwin 2.0.
\textbf{Act.}: Action-Only Denoising.
\textbf{App.}: Appearance-Only Foresight.
\textbf{ST}: Full Spatial-Temporal Foresight.
\textbf{Dis./En.}: GASAM disabled/enabled.
}
\label{tab:ablation_detailed}
\resizebox{\textwidth}{!}{
\begin{tabular}{lcccccccccccc}
\toprule
\multirow{3}{*}{Task}
& \multicolumn{6}{c}{Randomized}
& \multicolumn{6}{c}{Clean} \\
\cmidrule(lr){2-7}\cmidrule(lr){8-13}
& \multicolumn{2}{c}{Act.}
& \multicolumn{2}{c}{App.}
& \multicolumn{2}{c}{ST}
& \multicolumn{2}{c}{Act.}
& \multicolumn{2}{c}{App.}
& \multicolumn{2}{c}{ST} \\
\cmidrule(lr){2-3}\cmidrule(lr){4-5}\cmidrule(lr){6-7}
\cmidrule(lr){8-9}\cmidrule(lr){10-11}\cmidrule(lr){12-13}
& Dis. & En. & Dis. & En. & Dis. & En. & Dis. & En. & Dis. & En. & Dis. & En. \\
\midrule
Adjust Bottle & 91\% & 90\% & 91\% & 100\% & 100\% & 97\% & 96\% & 98\% & 99\% & 97\% & 97\% & 98\% \\
Beat Block Hammer & 59\% & 77\% & 62\% & 88\% & 90\% & 89\% & 49\% & 64\% & 75\% & 88\% & 88\% & 92\% \\
Blocks Ranking Rgb & 11\% & 23\% & 95\% & 94\% & 95\% & 98\% & 13\% & 12\% & 91\% & 96\% & 98\% & 100\% \\
Blocks Ranking Size & 9\% & 60\% & 66\% & 66\% & 71\% & 74\% & 6\% & 61\% & 70\% & 64\% & 74\% & 71\% \\
Click Alarmclock & 100\% & 98\% & 100\% & 100\% & 100\% & 100\% & 100\% & 100\% & 99\% & 100\% & 100\% & 100\% \\
Click Bell & 100\% & 99\% & 100\% & 97\% & 99\% & 100\% & 100\% & 100\% & 100\% & 100\% & 100\% & 100\% \\
Dump Bin Bigbin & 63\% & 87\% & 89\% & 85\% & 94\% & 97\% & 69\% & 90\% & 94\% & 89\% & 94\% & 93\% \\
Grab Roller & 96\% & 100\% & 99\% & 100\% & 100\% & 100\% & 99\% & 98\% & 100\% & 100\% & 100\% & 100\% \\
Handover Block & 8\% & 28\% & 32\% & 32\% & 28\% & 48\% & 15\% & 13\% & 21\% & 31\% & 24\% & 56\% \\
Handover Mic & 78\% & 86\% & 69\% & 97\% & 99\% & 99\% & 75\% & 99\% & 95\% & 100\% & 99\% & 100\% \\
Hanging Mug & 11\% & 8\% & 53\% & 50\% & 51\% & 72\% & 6\% & 10\% & 43\% & 42\% & 68\% & 69\% \\
Lift Pot & 86\% & 97\% & 98\% & 100\% & 100\% & 100\% & 85\% & 90\% & 100\% & 100\% & 100\% & 100\% \\
Move Can Pot & 66\% & 85\% & 92\% & 69\% & 95\% & 98\% & 54\% & 61\% & 78\% & 91\% & 96\% & 100\% \\
Move Pillbottle Pad & 89\% & 95\% & 97\% & 98\% & 100\% & 100\% & 85\% & 72\% & 97\% & 100\% & 99\% & 100\% \\
Move Playingcard Away & 95\% & 96\% & 100\% & 100\% & 97\% & 100\% & 95\% & 96\% & 98\% & 98\% & 86\% & 100\% \\
Move Stapler Pad & 32\% & 80\% & 60\% & 70\% & 75\% & 93\% & 31\% & 71\% & 73\% & 87\% & 87\% & 93\% \\
Open Laptop & 90\% & 94\% & 94\% & 87\% & 87\% & 91\% & 89\% & 94\% & 94\% & 93\% & 90\% & 93\% \\
Open Microwave & 39\% & 10\% & 76\% & 20\% & 19\% & 61\% & 28\% & 43\% & 68\% & 29\% & 34\% & 58\% \\
Pick Diverse Bottles & 78\% & 17\% & 89\% & 92\% & 95\% & 96\% & 90\% & 38\% & 80\% & 90\% & 95\% & 98\% \\
Pick Dual Bottles & 89\% & 15\% & 95\% & 100\% & 99\% & 100\% & 96\% & 21\% & 99\% & 99\% & 100\% & 100\% \\
Place A2b Left & 75\% & 87\% & 89\% & 82\% & 84\% & 91\% & 71\% & 83\% & 79\% & 91\% & 90\% & 96\% \\
Place A2b Right & 71\% & 84\% & 91\% & 85\% & 92\% & 96\% & 69\% & 76\% & 93\% & 88\% & 95\% & 94\% \\
Place Bread Basket & 74\% & 86\% & 88\% & 96\% & 93\% & 99\% & 78\% & 74\% & 94\% & 96\% & 95\% & 100\% \\
Place Bread Skillet & 65\% & 87\% & 81\% & 93\% & 88\% & 95\% & 70\% & 71\% & 87\% & 88\% & 93\% & 96\% \\
Place Burger Fries & 99\% & 99\% & 89\% & 93\% & 93\% & 99\% & 95\% & 97\% & 100\% & 92\% & 96\% & 99\% \\
Place Can Basket & 24\% & 65\% & 79\% & 72\% & 78\% & 88\% & 20\% & 44\% & 88\% & 86\% & 84\% & 89\% \\
Place Cans Plasticbox & 87\% & 95\% & 95\% & 91\% & 100\% & 100\% & 82\% & 65\% & 99\% & 99\% & 100\% & 100\% \\
Place Container Plate & 94\% & 100\% & 98\% & 97\% & 100\% & 100\% & 93\% & 98\% & 99\% & 100\% & 100\% & 100\% \\
Place Dual Shoes & 94\% & 92\% & 72\% & 98\% & 97\% & 94\% & 58\% & 85\% & 80\% & 75\% & 94\% & 96\% \\
Place Empty Cup & 96\% & 100\% & 99\% & 99\% & 99\% & 100\% & 94\% & 95\% & 98\% & 100\% & 100\% & 100\% \\
Place Fan & 63\% & 86\% & 88\% & 89\% & 87\% & 92\% & 60\% & 87\% & 89\% & 85\% & 84\% & 95\% \\
Place Mouse Pad & 35\% & 80\% & 82\% & 83\% & 86\% & 98\% & 39\% & 70\% & 77\% & 79\% & 94\% & 96\% \\
Place Object Basket & 17\% & 51\% & 87\% & 89\% & 91\% & 91\% & 23\% & 57\% & 90\% & 84\% & 91\% & 89\% \\
Place Object Scale & 61\% & 86\% & 84\% & 90\% & 87\% & 88\% & 64\% & 83\% & 87\% & 90\% & 88\% & 89\% \\
Place Object Stand & 98\% & 97\% & 99\% & 100\% & 99\% & 100\% & 93\% & 97\% & 96\% & 98\% & 100\% & 100\% \\
Place Phone Stand & 67\% & 81\% & 87\% & 90\% & 93\% & 97\% & 67\% & 73\% & 84\% & 88\% & 93\% & 91\% \\
Place Shoe & 91\% & 97\% & 100\% & 99\% & 95\% & 99\% & 88\% & 96\% & 100\% & 99\% & 99\% & 100\% \\
Press Stapler & 91\% & 98\% & 100\% & 96\% & 95\% & 100\% & 84\% & 99\% & 99\% & 99\% & 97\% & 100\% \\
Put Bottles Dustbin & 7\% & 31\% & 73\% & 89\% & 91\% & 93\% & 5\% & 30\% & 82\% & 86\% & 92\% & 96\% \\
Put Object Cabinet & 19\% & 32\% & 46\% & 74\% & 78\% & 77\% & 7\% & 31\% & 41\% & 71\% & 75\% & 84\% \\
Rotate Qrcode & 57\% & 71\% & 80\% & 77\% & 79\% & 84\% & 72\% & 78\% & 83\% & 70\% & 84\% & 94\% \\
Scan Object & 52\% & 81\% & 87\% & 87\% & 96\% & 96\% & 58\% & 70\% & 85\% & 87\% & 90\% & 93\% \\
Shake Bottle & 97\% & 96\% & 98\% & 94\% & 97\% & 97\% & 99\% & 100\% & 99\% & 98\% & 96\% & 100\% \\
Shake Bottle Horizontally & 94\% & 97\% & 97\% & 97\% & 96\% & 99\% & 99\% & 100\% & 96\% & 98\% & 97\% & 100\% \\
Stack Blocks Three & 28\% & 94\% & 95\% & 96\% & 99\% & 100\% & 23\% & 90\% & 91\% & 96\% & 99\% & 97\% \\
Stack Blocks Two & 73\% & 87\% & 100\% & 100\% & 98\% & 100\% & 65\% & 55\% & 94\% & 100\% & 100\% & 100\% \\
Stack Bowls Three & 20\% & 42\% & 87\% & 86\% & 85\% & 92\% & 17\% & 37\% & 88\% & 90\% & 92\% & 91\% \\
Stack Bowls Two & 74\% & 98\% & 92\% & 99\% & 93\% & 99\% & 66\% & 95\% & 97\% & 96\% & 98\% & 100\% \\
Stamp Seal & 85\% & 91\% & 95\% & 99\% & 98\% & 99\% & 80\% & 93\% & 99\% & 99\% & 98\% & 100\% \\
Turn Switch & 50\% & 58\% & 75\% & 63\% & 80\% & 89\% & 51\% & 65\% & 64\% & 71\% & 77\% & 85\% \\
\midrule
Average (\%) & 64.96\% & 75.88\% & 85.80\% & 86.96\% & 88.82\% & 93.30\% & 63.42\% & 72.50\% & 86.64\% & 87.86\% & 90.40\% & 93.82\% \\
\bottomrule
\end{tabular}
}
\end{table*}

\section{Limitations}
\label{app:limitations}
Although STARRY improves spatial-temporal action generation in both simulation and real-world experiments, several limitations remain. First, real-world evaluation is conducted on a limited number of bimanual manipulation tasks, and broader validation across more robot embodiments and open-ended environments is needed. Second, GASAM benefits from accurate predicted depth and end-effector geometry; under highly ambiguous observations, less reliable geometry prediction may reduce the effectiveness of the modulation signal. Finally, STARRY builds on large video and vision-language backbones, which increases training cost compared with smaller reactive policies.

\section{Impact statement}
\label{app:impact}
STARRY aims to improve robotic manipulation by enhancing spatial-temporal prediction and geometry-aware action generation. This may benefit applications such as assistive robotics, manufacturing, and household automation by improving execution reliability in spatially demanding tasks. Potential risks include unsafe robot behaviors under distribution shift or inappropriate deployment in unstructured environments. Therefore, STARRY should be viewed as a research-stage system and deployed only with appropriate physical safety constraints, monitoring, and human oversight.


\end{document}